\documentclass{article}

\usepackage[table]{xcolor}
\usepackage{microtype}
\usepackage{graphicx}
\usepackage{subfigure}
\usepackage{booktabs} 
\usepackage{authblk}
\usepackage{hyperref}

\usepackage[accepted]{icml2024}

\usepackage{amsmath}
\usepackage{amssymb}
\usepackage{mathtools}
\usepackage{amsthm}

\usepackage[capitalize,noabbrev]{cleveref}

\theoremstyle{plain}

\theoremstyle{definition}

\theoremstyle{remark}

\usepackage[textsize=tiny]{todonotes}
\usepackage{multirow}

\usepackage{graphicx}
\usepackage{inconsolata}
\usepackage{microtype}
\usepackage{epigraph}
\usepackage{graphicx}
\usepackage{microtype}
\usepackage{tikz}
\usepackage{pgfplots}

\usepackage{tabularx}
\usepackage{subcaption}
\usepackage[edges]{forest}
\definecolor{paired-light-blue}{RGB}{198, 219, 239}
\definecolor{paired-dark-blue}{RGB}{49, 130, 188}
\definecolor{paired-light-orange}{RGB}{251, 208, 162}
\definecolor{paired-dark-orange}{RGB}{230, 85, 12}
\definecolor{paired-light-green}{RGB}{199, 233, 193}
\definecolor{paired-dark-green}{RGB}{49, 163, 83}
\definecolor{paired-light-purple}{RGB}{218, 218, 235}
\definecolor{paired-dark-purple}{RGB}{117, 107, 176}
\definecolor{paired-light-gray}{RGB}{217, 217, 217}
\definecolor{paired-dark-gray}{RGB}{99, 99, 99}
\definecolor{paired-light-pink}{RGB}{222, 158, 214}
\definecolor{paired-dark-pink}{RGB}{123, 65, 115}
\definecolor{paired-light-red}{RGB}{231, 150, 156}
\definecolor{paired-dark-red}{RGB}{131, 60, 56}
\definecolor{paired-light-yellow}{RGB}{231, 204, 149}
\definecolor{paired-dark-yellow}{RGB}{141, 109, 49}
\tikzset{%
    parent/.style =          {align=center,text width=2.8cm,rounded corners=3pt, line width=0.3mm, fill=gray!10,draw=gray!80},
    child/.style =           {align=center,text width=2.3cm,rounded corners=3pt, fill=blue!10,draw=blue!80,line width=0.3mm},
    grandchild/.style =      {align=center,text width=2cm,rounded corners=3pt},
    greatgrandchild/.style = {align=center,text width=1.5cm,rounded corners=3pt},
    greatgrandchild2/.style = {align=center,text width=1.5cm,rounded corners=3pt},    
    referenceblock/.style =  {align=center,text width=1.5cm,rounded corners=2pt},
    acquisition/.style =    {align=center,text width=2.2cm,rounded corners=3pt, fill=paired-light-blue!50,draw=paired-dark-blue!65,line width=0.3mm},   
    acquisition_work/.style =           {align=center, text width=8.9cm,rounded corners=3pt, fill=paired-light-blue!50,draw=blue!0,line width=0.3mm},  
    representation/.style =           {align=center,text width=2.2cm,rounded corners=3pt, fill=paired-light-orange!50,draw=paired-dark-orange!65,line width=0.3mm},   
    representation_work/.style =           {align=center,text width=6cm,rounded corners=3pt, fill=paired-light-orange!50,draw=red!0,line width=0.3mm},
    representation_work_2/.style =           {align=center,text width=8.7cm,rounded corners=3pt, fill=paired-light-orange!50,draw=red!0,line width=0.3mm},
    probing/.style =           {align=center,text width=2.2cm,rounded corners=3pt, fill= paired-light-green!50,draw=paired-dark-green!75,line width=0.3mm},   
    probing_work/.style =           {align=center,text width=6cm,rounded corners=3pt, fill= paired-light-green!50,draw= cyan!0,line width=0.3mm},    
    cus_probing_work/.style =           {align=center,text width=8.7cm,rounded corners=3pt, fill= paired-light-green!50,draw= cyan!0,line width=0.3mm},
    editing/.style =           {align=center,text width=2.2cm,rounded corners=3pt, fill= paired-light-purple!50,draw=paired-dark-purple!75,line width=0.3mm},   
    editing_work/.style =           {align=center,text width=8.7cm,rounded corners=3pt, fill= paired-light-purple!50,draw= orange!0,line width=0.3mm},        
    application/.style =           {align=center,text width=2.2cm,rounded corners=3pt, fill= paired-light-red!35,draw=paired-light-red!90,line width=0.3mm},   
    application_work/.style =       {align=center,text width=8.7cm,rounded corners=3pt, fill= paired-light-red!35,draw= magenta!0,line width=0.3mm},         
}

\icmltitlerunning{Towards Next Generation Post-training Paradigm of Foundation Models via Verifier Engineering}

\begin{document}

\twocolumn[
\icmltitle{Search, Verify and Feedback: Towards Next Generation \\ Post-training Paradigm of Foundation Models via Verifier Engineering}


\icmlsetsymbol{equal}{*}

\begin{icmlauthorlist}
\icmlauthor{Xinyan Guan\textsuperscript{*}}{}
\icmlauthor{Yanjiang Liu\textsuperscript{*}}{}
\icmlauthor{Xinyu Lu\textsuperscript{*}}{}
\icmlauthor{Boxi Cao\textsuperscript{}}{}
\icmlauthor{Ben He\textsuperscript{}}{}
\icmlauthor{Xianpei Han\textsuperscript{}}{}
\icmlauthor{Le Sun\textsuperscript{}}{}\\
\icmlauthor{Jie Lou\textsuperscript{†}}{}
\icmlauthor{Bowen Yu\textsuperscript{†}}{}
\icmlauthor{Yaojie Lu\textsuperscript{†}}{}
\icmlauthor{Hongyu Lin\textsuperscript{†}}{}
\end{icmlauthorlist}

\begin{center}
Chinese Information Processing Laboratory, \\Institute of Software, Chinese Academy of Sciences \\

E-mail to: hongyu@iscas.ac.cn

\url{https://github.com/icip-cas/Verifier-Engineering}
\end{center}

\icmlkeywords{Machine Learning, ICML}

\vskip 0.3in

]




\let\thefootnote\relax\footnote{\textsuperscript{*}Equal contribution.}
\let\thefootnote\relax\footnote{\textsuperscript{†}Corresponding authors. Bowen Yu is affiliated with Qwen Team, Alibaba Group. Jie Lou is affiliated with Xiaohongshu Inc. This work is initiated by 4 corresponding authors in a personal meeting.}

\begin{abstract}

The evolution of machine learning has increasingly prioritized the development of powerful models and more scalable supervision signals.
However, the emergence of foundation models presents significant challenges in providing effective supervision signals necessary for further enhancing their capabilities. 
Consequently, there is an urgent need to explore novel supervision signals and technical approaches. 
In this paper, we propose \textit{\textbf{verifier engineering}}, a novel post-training paradigm specifically designed for the era of foundation models. 
The core of verifier engineering involves leveraging a suite of automated verifiers to perform verification tasks and deliver meaningful feedback to foundation models.
We systematically categorize the verifier engineering process into three essential stages: search, verify, and feedback, and provide a comprehensive review of state-of-the-art research developments within each stage.
We believe that verifier engineering constitutes a fundamental pathway toward achieving Artificial General Intelligence.

\end{abstract}
\section{Introduction}

The evolution of machine learning~\cite{jordan2015machine} has undergone a progressive journey characterized by the pursuit of increasingly powerful models and the scaling of supervision signals~\cite{friedland2018capacityscalinglawartificial,cao2024scalableautomatedalignmentllms,Sevilla_2022}. Over the past decades, both model capacity and the scale of supervision signals have escalated in a synergistic manner. Powerful models necessitate more scalable and effective supervision signals to fully utilize their parameters. Conversely, the expansion of supervision signals requires models with enhanced capacities to effectively exploit these signals, thereby achieving more generalized capabilities.

In the early stages of machine learning, models were constrained by their limited capacity. This period was characterized as \textit{\textbf{feature engineering}}, during which domain experts manually designed and extracted relevant features. 
Classical algorithms, such as Support Vector Machines~\cite{hearst1998support} and Decision Trees~\cite{quinlan1986induction}, relied heavily on meticulously crafted features due to their architectural limitations. 
These algorithms achieved optimal performance through carefully designed feature extraction techniques, with Term Frequency-Inverse Document Frequency~\cite{robertson2009probabilistic} serving as a prominent example.

\begin{table}[t]
\centering
\resizebox{0.48\textwidth}{!}{
\begin{tabular}{p{2.0cm}p{3.0cm}p{3.2cm}p{3.0cm}}
\toprule
 & \textbf{Feature} \textbf{Engineering} & \textbf{Data}  \textbf{Engineering} & \textbf{Verifier}  \textbf{Engineering} \\
\midrule
\textbf{Representative Models} & Machine Learning\newline Models & Deep Learning\newline Models & Foundation Models \\
\footnotesize e.g. & \footnotesize SVM, XGBoost & \footnotesize CNN, LSTM & \footnotesize LLMs, VLMs \\
\midrule
\textbf{Supervision} & Manual Features & Human Annotations & Verifier Feedback \\
\midrule
\textbf{Scope} & Task-Specific & Multiple Related Tasks & General Intelligence \\
\midrule
\textbf{Generalization} & Limited & Relatively high & High \\
\midrule
\textbf{Scalability} & Limited & Moderate & High \\
\bottomrule
\end{tabular}}
\caption{Comparison of feature engineering, data engineering, and verifier engineering}
\label{tab:compare}
\end{table}

\begin{figure*}
    \centering
    \includegraphics[width=0.9\linewidth]{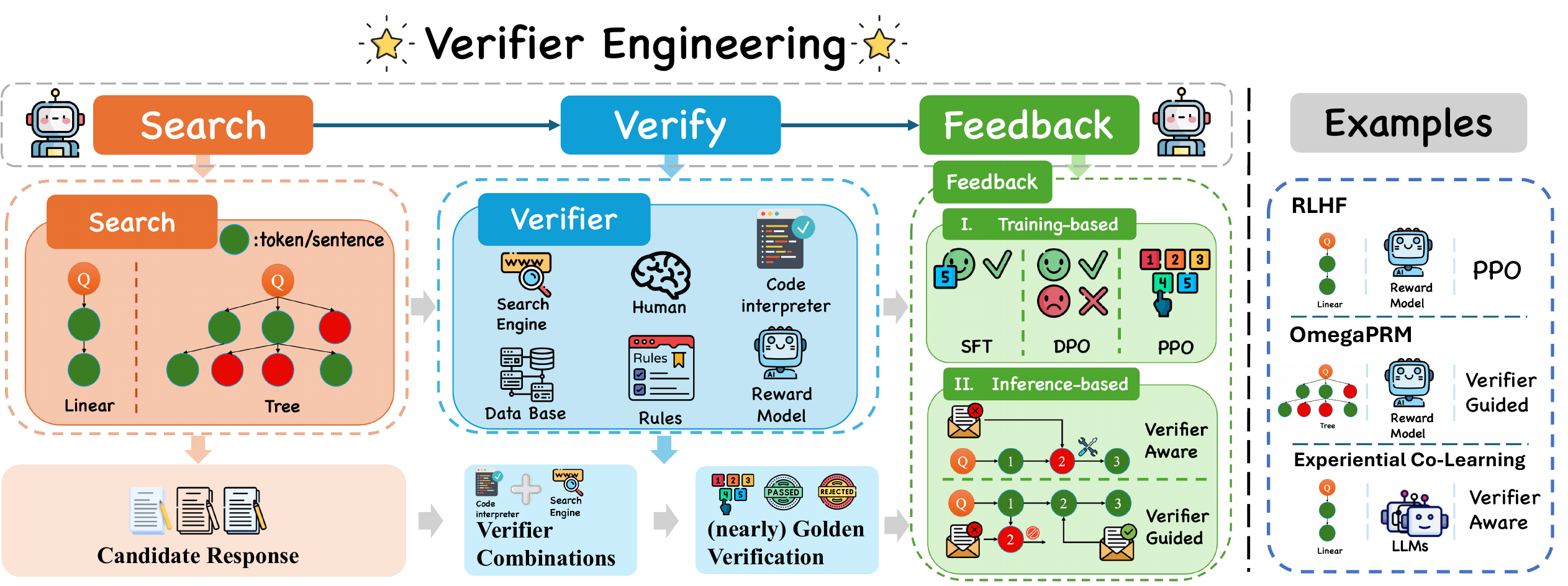}
    \caption{\textbf{Framework of \textit{verifier engineering}}: The fundamental stages of verifier engineering include Search, Verify, and Feedback. Given an instruction, the process begins with generating candidate responses (Search), followed by evaluating these candidates using appropriate verifier combinations (Verify), and concludes with optimizing the model's output distribution (Feedback). This framework can explain various approaches, from training-based methods like RLHF~\cite{ouyang2022training} to inference-based techniques such as OmegaPRM~\cite{luo2024improve} and Experiential Co-Learning~\cite{qian2023experiential}. We systematically categorize existing approaches into these three stages in Table~\ref{tab:exist-work}.}
    \label{fig:framework}
\end{figure*}

However, as addressed increasingly complex problems, the limitations of manual feature engineering became more pronounced, underscoring the need for more scalable approaches to construct features. 
The emergence of deep learning~\cite{SCHMIDHUBER201585} approximately two decades ago marked a transformative shift, inaugurating the \textit{\textbf{data engineering}} era. 
This new paradigm represented a fundamental departure from handcrafted features, emphasizing instead the curation of high-quality datasets and annotations to facilitate automated knowledge acquisition and pattern recognition across diverse domains and tasks. 
The remarkable success of landmark projects such as ImageNet~\cite{deng2009imagenet} and BERT~\cite{brown2020language} validated the effectiveness of this data-centric approach.

Unfortunately, the emergence of foundation models in recent years~\cite{ouyang2022training,touvron2023llama,BetkerImprovingIG,dosovitskiy2021imageworth16x16words} has made it increasingly difficult to enhance model capabilities solely through data engineering. Specifically, foundation models, particularly large language models (LLMs)~\cite{ouyang2022training,touvron2023llama}, have demonstrated extraordinary abilities, increasingly matching or surpassing human performance across various domains. Nevertheless, the traditional data engineering approach of augmenting these models through large-scale data construction has reached its practical limitations. This limitation is evident in two primary challenges: the difficulty and unsustainable costs associated with high-quality human annotations for post-training~\cite{anthropic2024measuring,burns2023weaktostrong}, and the complexity involved in providing meaningful guidance that can further enhance model performance~\cite{wen2024language}. Consequently, the central challenge in the current era is determining how to supply more effective supervision signals to foundation models to achieve general artificial intelligence capabilities.

\begin{figure*}[!tp]
\centering
\resizebox{!}{\textwidth}{
    \begin{forest}
        for tree={
            forked edges,
            grow'=0,
            draw,
            rounded corners,
            node options={align=center,},
            text width=2.7cm,
            s sep=6pt,
            l sep=30pt,
            calign=edge midpoint,
        },
        [Verifier\\ Engineering, fill=gray!45, parent
            [Search \S\ref{sec:search}, for tree={acquisition}
                [Linear Search, 
                    [Speculative Decoding \citep{leviathan2023fastinferencetransformersspeculative};  Reject Sampling \cite{yuan2023scalingrelationshiplearningmathematical}; CoT \cite{wei2022chain};
                    etc., acquisition_work
                    ]
                ]
                [Tree Search,
                    [Beam Search \citep{vijayakumar2018diversebeamsearchdecoding}; 
                    MCTS ;ToT \citep{yao2024tree}; PoT \citep{luo2024pythonbestchoiceembracing}; CoT-SC \citep{wang2023selfconsistency};  etc., acquisition_work
                    ]
                ]
            ]
            [Verify \S\ref{sec:verify}, for tree={representation}
                [ Verification Form,
                    [Ranking, representation
                        [CAI~\citep{bai2022constitutional}; PairRM~\citep{jiang-etal-2023-llm}; etc., representation_work]
                    ]
                    [Binary, representation
                        [Code Interpreter~\citep{gao2023pal}; Game Scoring System~\citep{tsai2023can}; etc., representation_work]
                    ]
                    [Score, representation
                        [Bradley-Terry RM ~\citep{christiano2017deep,askell2021general,bai2022training}; Hinge RM~\citep{mu2024rule}; Focal RM~\citep{cai2024internlm2}; Cross-Entropy RM~\citep{cobbe2021training,wang2024mathshepherd} etc., representation_work]
                    ]
                    [Text, representation
                        [LLM-as-a-Judge~\citep{zheng2024judging}; Calculator; Browser~\citep{nakano2021webgpt}; Self-Critique~\citep{saunders2022self}; Critique-of-Critique~\citep{sun2024critique} ; etc., representation_work]
                    ]
                ]
                [ Verify Granularity,
                    [Token-level, representation
                    [Value Model~\citep{chen2024alphamath}; etc., representation_work]
                    ]
                    [Thought-level, representation
                        [PRM~\citep{lightman2023lets}; etc., representation_work]
                    ]
                    [Trajectory-level, representation
                        [ORM~\citep{cobbe2021training}; etc., representation_work]
                    ]
                ]
                [ Verifier Source, 
                    [Self-Supervised, representation
                        [Self-Rewarding~\citep{yuan2024self, wu2024meta};
                        Self-Critique~\citep{saunders2022self}; Quiet-STaR~\citep{zelikman2024quiet}; etc., representation_work]
                    ]
                    [Other-Supervised, representation
                        [Human~\citep{christiano2017deep}; Teacher Model~\citep{wen2024languagemodelslearnmislead}; etc., representation_work]
                    ]
                ]
            ]
            [Feedback \S\ref{sec:feedback}, for tree={probing}
                [Training Based,
                    [Imitation Learning ,  probing
                        [ SFT~\cite{ouyang2022training}; KD~\cite{hinton2015distilling}; etc., probing_work]
                    ]
                    [Preference Learning,  probing
                        [DPO \cite{rafailov2024direct}; IPO \cite{azar2024general}; KTO~\cite{ethayarajh2024kto} etc., probing_work]
                    ]
                    [Reinforcement Learning,  probing
                        [PPO~\cite{schulman2017proximal}; PPO-max~\cite{zheng2023secrets}; RLMEC~\cite{chen2024improving}  etc., probing_work]
                    ]             
                ]
                [Inference Based, 
                    [Verifier Guided ,  probing
                        [PRM \cite{lightman2023let}; BON \cite{sun2024fastbestofndecodingspeculative};  RAIN \cite{li2024rain};  etc., probing_work]
                    ]
                    [Verifier Aware,  probing
                    [Self-Debug \cite{chen2023teaching};  ReAct \cite{yao2022react}; Experiential Co-Learning \cite{qian2023experiential}; etc., probing_work]
                    ]
                ]
            ]
        ]
    \end{forest}
}
    \caption{Overview of verifier engineering methodologies, categorized into three main stages: Search, Verify, and Feedback. Each stage is further broken down into specific approaches, with references to notable works in each area.}
    \label{fig:tax}
\end{figure*}
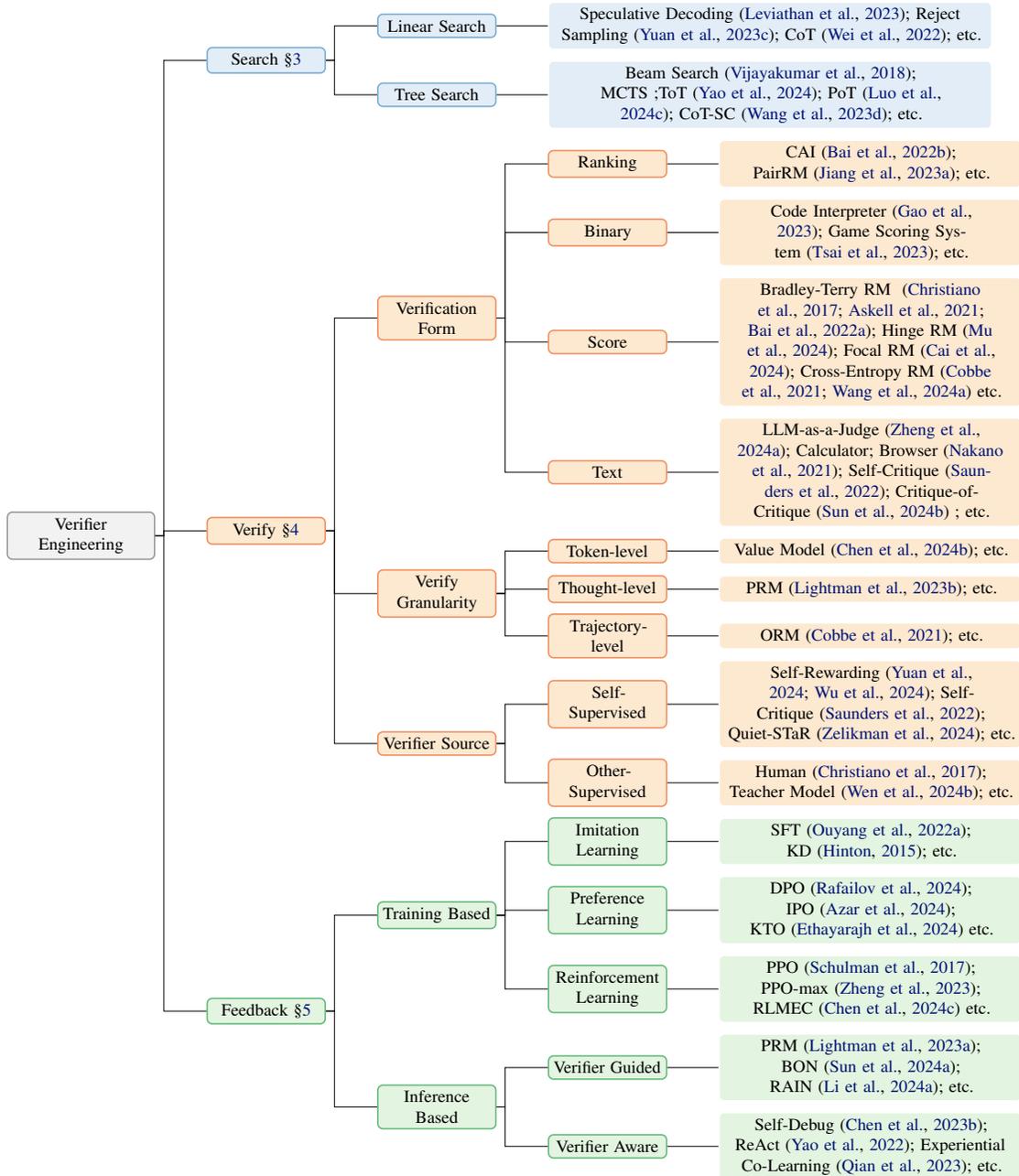

In this paper, we propose \textit{\textbf{verifier engineering}}, a novel post-training paradigm designed for the foundation model era. 
The essence of verifier engineering lies in extending the construction of supervision signals beyond traditional manual feature extraction and data annotation. 
Instead, it utilizes a suite of effective automated verifiers to perform verification tasks and provide meaningful feedback to foundation models. 
Table~\ref{tab:compare} delineates the key distinctions among feature engineering, data engineering, and verifier engineering.
This progression from \textbf{annotate and learn} to \textbf{search and verify} signifies a fundamental advancement in enhancing the capabilities of foundation models. 
Compared to preceding paradigms, verifier engineering streamlines the creation of verifiers and facilitates efficient feedback to foundation models through automated verification processes. 
Specifically, given an instruction, verifier engineering initiates by generating candidate responses, subsequently verifying these candidates using appropriate combinations of verifiers, and ultimately optimizes the model's output distribution.
Unlike existing methodologies such as Reinforcement Learning from Human Feedback (RLHF)~\cite{stiennon2020learning,ouyang2022training}, which depend on a limited set of verifier sources and feedback mechanisms, verifier engineering integrates multiple diverse verifiers to deliver more accurate and generalizable feedback signals. 
By shifting the improvement of foundation models from a data-centric endeavor to a systematic engineering challenge, verifier engineering emphasizes the design and orchestration of complex verification systems to ensure effective and efficient feedback.
Analogous to how feature engineering and data engineering achieved scalability in their respective eras, we posit that verifier engineering represents a crucial step toward the advancement of general artificial intelligence.

To this end, this paper provides a comprehensive exploration of the verifier engineering landscape,  breaking it down into three core stages: search, verify, and feedback. These stages are defined as follows:

\begin{enumerate}
    \item \textbf{Search}: Sampling representative or potentially problematic samples from model output distribution to reveal performance boundaries and limitations.
    \item \textbf{Verify}: Utilizing various verifiers to provide verification results on candidate responses, which may include evaluation metrics, rule detection, or selective manual annotation.
    \item \textbf{Feedback}: Leveraging verification results to enhance model performance through methods such as Supervised Fine-Tuning or In-Context Learning.
\end{enumerate}

To showcase the latest state-of-the-art research developments in these three stages, we provide a comprehensive review in the remainder of this paper: Section 2 presents the formal definition and preliminaries of verifier engineering. Sections 3-5 detail the three fundamental stages: search, verify, and feedback. Section 6 discusses current trends and limitations, followed by concluding remarks in Section 7.

\section{Verifier Engineering}
In this section, we formalize verifier engineering as a Goal-Conditioned Markov Decision Process (GC-MDP)~\cite{pmlr-v37-schaul15, plappert2018multigoalreinforcementlearningchallenging, liu2022goalconditionedreinforcementlearningproblems}, enabling a unified and systematic perspective on the field of verifier engineering. Then, we introduce how the concepts of search, verify, and feedback correspond within this modeling framework, and analyze them through examples.
Furthermore, we provide a comprehensive overview of the verifier engineering landscape by categorizing existing post-training approaches into three core stages, as summarized in Table~\ref{tab:exist-work}.

\subsection{Preliminary}
While LLMs are typically trained to maximize generation likelihood given input, this objective alone cannot guarantee desired post-training capabilities. 
To bridge this gap, we formalize verifier engineering as a GC-MDP, denoted as tuple (\textbf{\(S\)}, \textbf{\(A\)}, \textbf{\(T\)}, \textbf{\(G\)}, \textbf{\(R_g\)}, \textbf{\(p_g\)}), where:

\textbf{State Space \( S \)} represents the model's state during interaction, including input context, internal states, and intermediate outputs.

\textbf{Action Space \( A \)} represents the possible token selections at each generation step: $$ A = \{a_1, a_2, \dots, a_N\} $$

\textbf{Transition Function  \(T\) } typically defines the probability distribution over the next states, given the current state  $s \in S$  and action  $a \in A$. Specifically, in large language models, the state transition is a deterministic function. That is, once the current state and selected action (the generated token) are given, the next state is fully determined. The search stage of the verifier
engineering thus can be regarded as exploration in the action-state space in the condition of \(T\).

\textbf{Goal Space \( G \)} represents the various goals related to model capabilities. Each goal \( g \in G \) corresponds to a specific model capability, such as code, math, writing, and so on. The goal space is multi-dimensional and can encompass various aspects of model capabilities.

\textbf{Goal Distribution  $p_g$ } is the probability distribution over goals from the goal space $G$ . It represents the likelihood of a specific goal  $g \in G$  being selected at any given time. This distribution can be learned from human feedback or other external signals.

\textbf{Reward Function \( R_g(s, a) \) or \(R_g(s')\)} represents the reward the model receives given the goal \( g \), when taking action \( a \) from state \( s \) or at the transformed state \(s'\). 
This reward thus reflects the verification result of verifier engineering for the specific goal \( g \).
For example, if the goal \( g \) is "fairness," the reward might be based on whether the generated text avoids bias.

The objective for improving model capabilities can be defined as a goal-conditioned policy $\pi : S \times G \times A \rightarrow [0, 1]$  that maximizes the expectation of the cumulative return over the goal distribution. 
\begin{equation}
    J(\pi)=\mathbb{E}_{\substack{a_t \sim \pi\left(\cdot \mid s_t, g\right), g \sim p_g \\ s_{t+1} \sim \mathcal{T}\left(\cdot \mid s_t, a_t\right)}}\left[\sum_t \gamma^t R_g\left(s_t, a_t\right)\right]
\end{equation}

The reward function can be decomposed into sub-functions for different capability dimensions: 
\begin{equation}
\begin{aligned}
    R_g(s, a) &= \mathcal{F} \left(R_{g,i}\left(s, a\right) \mid i \in \mathcal{S}_g \right)
\end{aligned}
\end{equation}
where $\mathcal{S}_g$ represents selected sub-functions for goal \(g\), and $\mathcal{F}$ combines their evaluations. Following Probably Approximately Correct Theory (PAC)~\cite{Vapnik2000}, even with imperfect sub-functions, we can achieve reliable overall evaluation by combining multiple weak verifiers.

\subsection{Verifier Engineering Overview}

Building upon the GC-MDP framework, we demonstrate how the three stages of verifier engineering—\textbf{Search}, \textbf{Verify}, and \textbf{Feedback}—naturally align with specific components of this formalism. This mapping provides both theoretical grounding and insights into how each stage contributes to optimizing the policy distribution $\pi$ toward desired goals.

As illustrated in Figure~\ref{fig:framework}, the search stage can be divided into linear search and tree search methods. The verify stage selects appropriate verifiers and invokes them to provide verification results based on the candidate responses. The feedback stage improve the model's output using either training-based methods or inference-based methods.
For example, RLHF utilizes linear search to generate batches of responses, employs a reward model as the verifier, and applies the Proximal Policy Optimization (PPO) \cite{schulman2017proximal} algorithm to optimize the model based on the reward model's verification result. 
OmegaPRM \cite{luo2024improvemathematicalreasoninglanguage} employs a process reward model as a verifier and searches for the best result based on the PRM by maximizing the process reward scores. 
Experimental Co-Learning \cite{qian2023experiential} collaborative verifiers through multiple LLMs, enhancing both verify and model performance through historical interaction data.

In the GC-MDP framework, the three stages manifest as follows: 
Search corresponds to action selection, where state $s$ represents the current context or partial sequence, and action $a$ denotes the next token$s$ to generate.
Verify maps to the reward function $R_g(s, a)$, evaluating how well the generated outputs align with the specified goal $g$.
Feedback relates to policy distribution $\pi$ optimization, maximizing the expected cumulative return $J(\pi)$ based on verification results.

\section{Search}

\label{sec:search}

Search aims to identify high-quality generated sequences that align with the intended goals, forming the foundation of verifier engineering, which is critical for evaluating and refining foundation models. However, it is impractical to exhaustively search the entire state-action space due to its exponential growth—driven by the large vocabulary size \(N\) and the maximum generation length \(T\). To address this challenge, efficient searching, which seeks to navigate this vast space by prioritizing diverse and goal-oriented exploration, plays a critical role in improving model's performance. 

In this section, we first introduce how to implement diverse search from the perspective of search structure and discuss additional methods for further enhancing search diversity after the search structure is determined.

\subsection{Search Structure}
Search structure denotes the framework or strategy used to navigate the state-action space, which significantly influences the effectiveness and efficiency of the search process. Currently, there are two widely adopted structures for implementing search: linear search and tree search. Linear search progresses sequentially, making it effective for tasks involving step-by-step actions, while tree search examines multiple paths at each decision point, making it well-suited for tasks requiring complex reasoning.

\begin{itemize}
    \item \textbf{Linear Search} is a widely used search method where the model starts from an initial state and proceeds step by step, selecting one token at a time until reaching the terminal state~\cite{brown2020languagemodelsfewshotlearners,wang2024chain}. 

    The key advantage of linear search is its low computational cost, which makes it efficient in scenarios where actions can be selected sequentially to progress toward a goal. However, a notable limitation is that if a sub-optimal action is chosen early in the process, it may be challenging to rectify the subsequent sequence. Thus, during the progress of verifier engineering, careful verification at each step is crucial to ensure the overall effectiveness of the generation path.

    \item \textbf{Tree Search} involves exploring multiple potential actions at each step of generation, allowing for a broader exploration of the state-action space. For instance, techniques like Beam Search and ToT \cite{yao2024tree} incorporate tree structures to enhance exploration, improving model performance in reasoning tasks. TouT \cite{mo2023treeuncertainthoughtsreasoning} introduces uncertainty measurement based on Monte Carlo dropout, providing a more accurate evaluation of intermediate reasoning processes.

    This approach significantly improves the likelihood of discovering global optimal solution, particularly in environments with  complex state spaces. By simultaneously considering multiple paths, tree search mitigates the risk of being locked into sub-optimal decisions made early on, making it more robust in guiding the model toward an optimal outcome. However, this increased exploration comes at a higher computational cost, making tree search more suitable for scenarios when the optimal path is challenging to identify. To make tree search effective, the model must be continuously verified and prioritize paths that align better with the goal conditions.
\end{itemize}

\subsection{Additional Enhancement}
While search structures provide the foundational framework for navigating the state-action space, further enhancement is also crucial to enhance search performance. These enhancements address challenges such as balancing exploration and exploitation, escaping local optima, and improving the diversity of generated results. The enhancement strategies can be broadly categorized into two approaches: adjusting exploration parameters and intervening in the original state.

\paragraph{Adjusting Exploration Parameters} 
Techniques such as Monte Carlo Tree Search (MCTS), Beam Search, and Reject Sampling focus on refining the exploration process by adjusting parameters like temperature, Top-k~\citep{fan2018hierarchicalneuralstorygeneration}, or Top-p~\cite{holtzman2020curiouscaseneuraltext}. The challenge lies in balancing the trade-off between generating diverse outputs and maintaining high-quality sequences. For example, increasing the temperature parameter promotes greater randomness, fostering diversity but potentially reducing coherence. 

\paragraph{Intervening in the Original State} Another enhancement approach involves modifying the initial state to guide the search process toward specific goals. Methods like Chain of Thought (CoT) \cite{wei2022chain}, Logical Inference via Neurosymbolic Computation (LINC) \cite{Olausson_2023}, and Program of Thoughts (PoT) \cite{chen2023programthoughtspromptingdisentangling} exemplify this strategy. These interventions address the challenge of overcoming biases in the default state distribution. CoT enhances reasoning by introducing intermediate steps, improving interpretability and depth in generated sequences. LINC uses logical scenarios to encourage more structured and goal-oriented outputs. Similarly, PoT provides programmatic examples that guide models toward systematic problem-solving, effectively expanding the scope of exploration beyond the original state distribution.

\section{Verify}

\label{sec:verify}

Due to the long delay and high cost associated with human feedback, we cannot directly employ human efforts to evaluate each candidate response sampled by the model during training~\citep{leike2018scalable}. Therefore, we employ verifiers as proxies for human supervision in the training of foundation models. The verifiers play a crucial role in search-verify-feedback pipeline, and the quality and robustness of the verifiers directly impact the performance of the downstream policy~\citep{wen2024rethinking}.

In the context of GC-MDP, verify is typically defined as using a verifier that provides verification results based on the current state and a predefined goal:
\begin{equation}
    F \leftarrow R_g(s_{t-1}, a_t)
\end{equation}

where  \(F\)  represents the verification results provided by the verifier.  \(g\)  denotes the predefined goal we aim to achieve (e.g., helpfulness, honesty). The state \( s \) is typically composed of two concatenated components: the user's query or input and the model’s output content $ \{a_1, …, a_t\}$.

In this section, we classify individual verifiers across several key dimensions and summarize the representative type of verifiers in Table~\ref{tab:verifiers}. 

\subsection{A Comprehensive Taxonomy of Verifiers}

\paragraph{Perspective of Verification Form} The verification result form of verifiers can be divided into four categories: binary feedback~\citep{gao2023pal}, score feedback~\citep{bai2022training}, ranking feedback~\citep{jiang-etal-2023-llm}, and text feedback~\citep{saunders2022self}. These categories represent an increasing gradient of information richness, providing more information to the optimization algorithm. For instance, classic Bradley–Terry reward models~\citep{Bradley1952RankAO} can provide continuous score feedback which simply indicates correctness, while text-based feedback from generative reward models~\citep{zhang2024generativeverifiersrewardmodeling} and critique models~\citep{sun2024critique} offer more detailed information, potentially including rationals for scores or critiques.

\paragraph{Perspective of Verify Granularity} The verify granularity of verifiers can be divided into three levels: token-level~\citep{chen2024alphamath}, thought-level~\citep{lightman2023let}, and trajectory-level~\citep{cobbe2021training}. These levels correspond to the scope at which the verifier engages with the model’s generation. Token-level verifiers focus on individual next-token predictions, thought-level verifiers assess entire thought steps or sentences, and trajectory-level verifiers evaluate the overall sequence of actions. 
Although most RLHF practices~\citep{ouyangTrainingLanguageModels2022b, bai2022training} currently rely on full-trajectory scoring, coarse-grained ratings are challenging to obtain accurately~\citep{wen2024languagemodelslearnmislead}, as they involve aggregating finer-grained verification. Generally, from the perspective of human annotators, assigning finer-grained scores is easier when the full trajectory is visible. From a machine learning perspective, fine-grained verification is preferable \citep{lightman2023let}, as it mitigates the risks of shortcut learning and bias associated with coarse-grained verification, thereby enhancing generalization. A credit-assignment mechanism~\citep{leike2018scalableagentalignmentreward} can bridge the gap between coarse-grained ratings and fine-grained verification.

\renewcommand{\arraystretch}{1.5}  

\begin{table*}[t]
\resizebox{\linewidth}{!}{
    \begin{tabular}{lllll}
    \toprule
    \textbf{Verifier Type} & \textbf{Verification Form} & \textbf{Verify Granularity} & \textbf{Verifier Source} & \multicolumn{1}{p{10.085em}}{\textbf{ Extra Training}} \\
    \midrule
    Golden Annotation & Binary/Text & Thought Step/Full Trajectory & Program Based & No \\
    Rule-based & Binary/Text & Thought Step/Full Trajectory & Program Based & No \\
    Code Interpreter & Binary/Score/Text & Token/Thought Step/Full Trajectory & Program Based & No \\
    ORM   & Binary/Score/Rank/Text & Full Trajectory & Model Based & Yes \\
    Language Model & Binary/Score/Rank/Text & Thought Step/Full Trajectory & Model Based & Yes \\
    Tool  & Binary/Score/Rank/Text & Token/Thought Step/Full Trajectory & Program Based & No \\
    Search Engine & Text  & Thought Step/Full Trajectory & Program Based & No \\
    PRM   & Score & Token/Thought Step & Model Based & Yes \\
    Knowledge Graph & Text  & Thought Step/Full Trajectory & Program Based & No \\
    \bottomrule
    \end{tabular}%
}
\caption{\textbf{A comprehensive taxonomy of verifiers across four dimensions: } verification form, verify granularity, verifier source, and the need for extra training.}
\label{tab:verifiers}
\end{table*}

\paragraph{Perspective of Verifier Source} Verifiers can be divided into program-based and model-based from the perspective of verifier source. Program-based verifiers provide deterministic verification, typically generated by predefined rules or logic embedded in fixed programs. These program-based verifiers offer consistent and interpretable evaluations but may lack flexibility when dealing with complex, dynamic environments. On the other hand, model-based verifiers rely on probabilistic models to generate verification results. These verifiers adapt to varying contexts and tasks through learning, allowing for more nuanced and context-sensitive evaluations. However, model-based verifiers introduce an element of uncertainty and can require significant training and computational resources to ensure accuracy and robustness.

\paragraph{Perspective of Extra Training} Verifiers can also be divided into two categories based on whether they require additional specialized training. Verifiers requiring additional training are typically fine-tuned on specific task-related data, allowing them to achieve higher accuracy in particular problem domains~\citep{markov2023holistic}. However, their performance can be heavily influenced by the distribution of the training data, potentially making them less generalizable to different contexts. On the other hand, verifiers that do not require extra training are often based on pre-existing models~\citep{zelikman2024quiet, zheng2024judging}. While they may not reach the same level of accuracy as their task-specific counterparts, they are generally more robust to variations in data distribution, making them less dependent on specific training sets. This trade-off between accuracy and data sensitivity is a key consideration when selecting a verifier for a given application.

\subsection{Constructing Verifiers across Tasks} 

Different tasks have varying requirements for verifiers, serving as critical tools to enhance the performance of foundation models. In this section, we will highlight key practices for constructing verifiers in representative fields, including safety and reasoning.

\paragraph{Safety} 

In safety-critical applications, verifiers play a crucial role in ensuring that foundation models adhere to ethical standards and avoid generating harmful or inappropriate content. Program-based verifiers can enforce strict guidelines by filtering out outputs that include prohibited content, such as hate speech or sensitive personal information. For instance, a content moderation system might employ predefined keywords and patterns to identify and block offensive language. However, a limitation of program-based approaches is their vulnerability to adversarial attacks, as paraphrased content can often bypass these filters~\citep{krishna-etal-2020-reformulating}. In contrast, model-based verifiers, such as toxicity classifiers~\citep{lees2022new, markov2023holistic, inan2023llama}, offer probabilistic assessments of content safety, enabling more nuanced evaluations. A middle-ground approach is rule-based reward models (RRMs)~\citep{mu2024rule}, which balance interpretability with generalization capabilities. Integrating verifiers into both the training and deployment phases allows foundation models to align more closely with safety requirements, reducing the likelihood of unintended or harmful outputs.
\vspace{-0.25cm}
\paragraph{Reasoning} 

In tasks requiring logical deduction or problem-solving, verifiers can assess the correctness and coherence of each reasoning step. Token-level verifiers offer fine-grained evaluations of individual tokens or symbols, which is particularly useful in mathematical computations and code generation~\citep{wang2023mathcoderseamlesscodeintegration}. Thought-level verifiers, on the other hand, evaluate entire sentences or reasoning steps to ensure that each component of the argument is both valid and logically consistent~\citep{lightman2023let, li-etal-2023-making, xie2024monte}. Trajectory-level verifiers assess the overall solution or proof, providing comprehensive verification results on the coherence of the model’s reasoning~\citep{cobbe2021training, yu-etal-2024-ovm, wang2024mathshepherd}. For instance, in mathematical theorem proving, a program-based verifier like Lean~\citep{Moura2015TheLT} can check each proof step's validity against formal logic rules~\citep{lin2024lean}, while a model-based verifier can assess the plausibility of reasoning steps through scores and natural language explanations~\citep{zhang2024generative}, offering critiques for further refinement~\citep{kumar2024traininglanguagemodelsselfcorrect}. A simpler yet widely-used approach involves using manually annotated correct answers as verifiers to filter model outputs, progressively enhancing model performance~\citep{zelikman2022star}.
\section{Feedback}
\label{sec:feedback}

After obtaining the verification result, we aim to enhance the capabilities of the foundation model, we define this process as the feedback stage.
In this paper, feedback specifically refers to enhancing the foundation model’s capabilities based on the verification results.
The feedback stage is critical, as the effectiveness of the feedback method directly determines whether the foundation model's capabilities can be appropriately enhanced in response to the verification results.

In this section, we explore how verifier engineering utilizes search algorithms and verifiers to feedback verification results on foundation models. 
To maximize the objective function $J(\pi)$, the distribution of the policy $\pi$ can be optimized by adjusting $s_t$ or the parameters of $\pi$. 
This leads to two distinct feedback approaches: training-based feedback and inference-based feedback.
Training-based feedback involves updating model parameters using data efficiently obtained through searching and verifying. 
In contrast, inference-based feedback modifies the output distribution by incorporating search and verification results as auxiliary information during inference, without altering the model parameters.

\subsection{Training-based Feedback}

We categorize the common training strategies for training-based feedback into three types, based on the nature and organization of the data utilized:

\paragraph{Imitation Learning} Imitation Learning typically uses high-quality data selected by verifiers, employing supervised fine-tuning objectives like cross-entropy or knowledge distillation training objectives like Kullback-Leibler divergence~\cite{hinton2015distilling} to optimize the model. 

Various approaches are employed to enhance specific capabilities of the foundation model through imitation learning. 
To enhance mathematical reasoning capabilities in LLMs, approaches like STaR~\cite{zelikman2022star} and RFT~\cite{yuan2023scaling} use rule-based verifiers to compare solution outcomes, while WizardMath~\cite{luo2023wizardmath}, MARIO~\cite{liao2024mario}, and MetaMath~\cite{yu2023metamath} leverage verifiers to select responses from advanced LLMs or human inputs. Other methods such as MAmmoTH~\cite{yue2023mammoth} and MathCoder~\cite{wang2023mathcoder} utilize programmatic tools to verify comprehensive and detailed solutions.
For coding capability improvement, various methods including Code Alpaca~\cite{codealpaca}, WizardCoder~\cite{luo2023wizardcoder}, WaveCoder~\cite{yu2023wavecoder}, and OpenCodeInterpreter~\cite{zheng2024opencodeinterpreter} construct code instruction-following datasets by distilling knowledge from advanced LLMs.
To enhance instruction-following abilities, approaches like LLaMA-GPT4~\cite{peng2023instruction}, Baize~\cite{xu2023baize}, and Ultrachat~\cite{ding2023enhancing} employ verifiers to distill responses from advanced LLMs for supervised fine-tuning. Other methods such as Deita~\cite{liu2024what} and MoDS~\cite{du2023mods} implement a pipeline of verifiers to check complexity, quality, and diversity before selecting suitable data for SFT.

\paragraph{Preference Learning} Preference Learning leverages verification results to construct pairwise comparison data and employs optimization methods like DPO~\cite{rafailov2024direct}, KTO~\cite{ethayarajh2024kto}, IPO~\cite{azar2024general}, and PRO~\cite{song2024preference}. Through this approach, models learn to align their outputs with verifier-provided preferences.

Various techniques are adopted to boost the foundation model's capabilities in specific areas through preference learning.
For mathematical reasoning enhancement, MCTS-DPO~\cite{xie2024monte} combines Monte Carlo Tree Search~\cite{coulom2006efficient,kocsis2006bandit} with preference learning to generate and learn from step-level pairwise comparisons in an iterative online manner.
For coding capability improvement, CodeUltraFeedback~\cite{weyssow2024codeultrafeedback} constructs pairwise training data by using LLM verifiers to rank code outputs, then applies preference learning algorithms to optimize the model's performance.
For instruction-following enhancement, Self-Rewarding~\cite{yuan2024self} enables models to generate their own verification results for creating pairwise comparison data, followed by iterative self-improvement using the DPO method.

\paragraph{Reinforcement Learning} Reinforcement Learning optimizes models using reward signals from verifiers. Through environmental interaction and policy updates using algorithms like PPO~\cite{schulman2017proximal}, PPO-max~\cite{zheng2023secrets}, models iteratively improve their generation quality.

Multiple approaches are used to enhance the foundation model’s capabilities in specific domains using reinforcement learning.
For mathematical reasoning enhancement, Math-Shepherd~\cite{wang2023math} implements step-wise reward mechanisms to guide progressive improvement in mathematical problem-solving capabilities.
For coding capability improvement, methods like RLTF~\cite{liu2023rltf} and PPOCoder~\cite{shojaee2023execution} leverage code execution results as reward signals to guide models toward more effective coding solutions.
For instruction-following enhancement, approaches like InstructGPT~\cite{ouyang2022training} and Llama~\cite{touvron2023llama,touvron2023llama2,dubey2024llama} employ reward models trained to evaluate response helpfulness, optimizing models for better instruction adherence.

\subsection{Inference-based Feedback}
In inference-based feedback, we modify inputs or inference strategies to obtain better outputs without changing model parameters. 
This approach is divided into two categories based on the visibility of verification results to the model: verifier-guided feedback and verifier-aware feedback.

\paragraph{Verifier-Guided} In verifier-guided feedback, verifiers evaluate and select the most appropriate outputs from model-generated content without direct model interaction.

For example, \citet{lightman2023let} and \citet{snell2024scaling} implement tree search algorithms guided by progress rewards, while ToT~\cite{yao2024tree} employs language model verifiers to direct its tree search process. In the realm of contrastive decoding~\cite{li2022contrastive,o2023contrastive}, advanced language models serve as token logits verifiers to optimize output distributions.

\paragraph{Verifier-Aware} Verifier-Aware feedback integrates verifier feedback directly into a model's operational context to enhance the content generation process. This approach allows the model to actively consider and incorporate verifier feedback while producing its output.

Various strategies are employed to enhance specific capabilities of foundation models through verifier-aware feedback.
For mathematical and coding enhancement, CRITIC~\cite{gou2023critic} utilizes feedback from calculators and code interpreters to refine solutions, while Self-debug~\cite{chen2023teaching} improves code quality through execution result analysis.
For hallucination mitigation, approaches like ReAct~\cite{yao2022react}, KGR~\cite{guan2024mitigating}, and CRITIC~\cite{gou2023critic} integrate continuous feedback from search engines and knowledge graphs to ensure factual accuracy. In a similar vein, Self-Refine~\cite{madaan2024self} employs language model verifiers to iteratively improve response quality.

\section{Discussion and Insights}
In this section, we provide a detailed examination of the insights derived from our framework. We begin by revisiting SFT, DPO, and RLHF through the lens of verifier engineering. 
Subsequently, we conduct an independent analysis of each stage within the framework.
Finally, we offer a systematic evaluation of potential challenges inherent to the framework as a whole.

\subsection{Revisiting SFT, DPO and RLHF from Verifier Engineering}

Our proposed approach to verifier engineering provides a unified perspective on commonly used post-training algorithms, offering valuable insights into their mechanisms.

\begin{figure}[h]
    \centering
    \begin{minipage}[b]{0.13\textwidth}
        \centering
        \includegraphics[width=\textwidth]{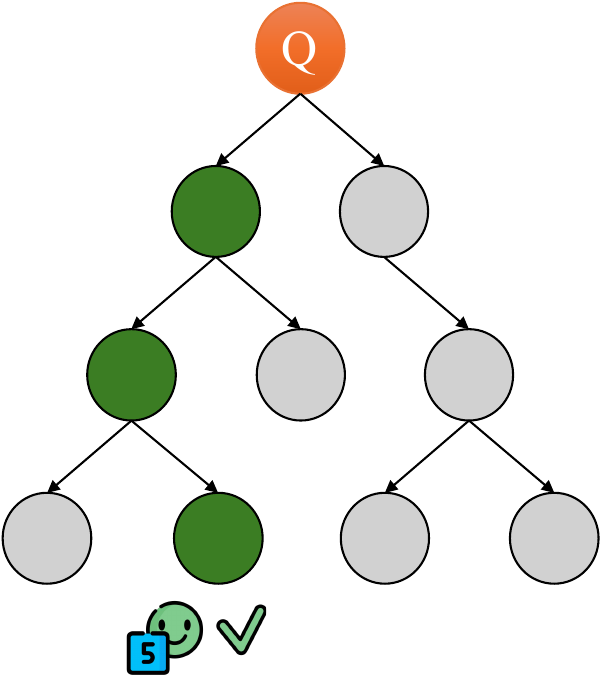}
        \vspace{0.5em}
        (a) SFT
    \end{minipage}
    \hspace{0.05\linewidth}
    \begin{minipage}[b]{0.13\textwidth}
        \centering
        \includegraphics[width=\textwidth]{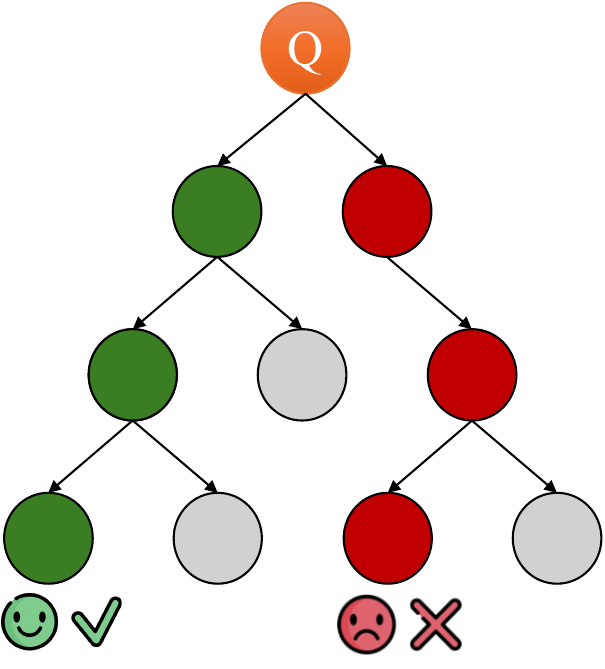}
        \vspace{0.5em}
        (b) DPO
    \end{minipage}
    \hspace{0.05\linewidth}
    \begin{minipage}[b]{0.13\textwidth}
        \centering
        \includegraphics[width=\textwidth]{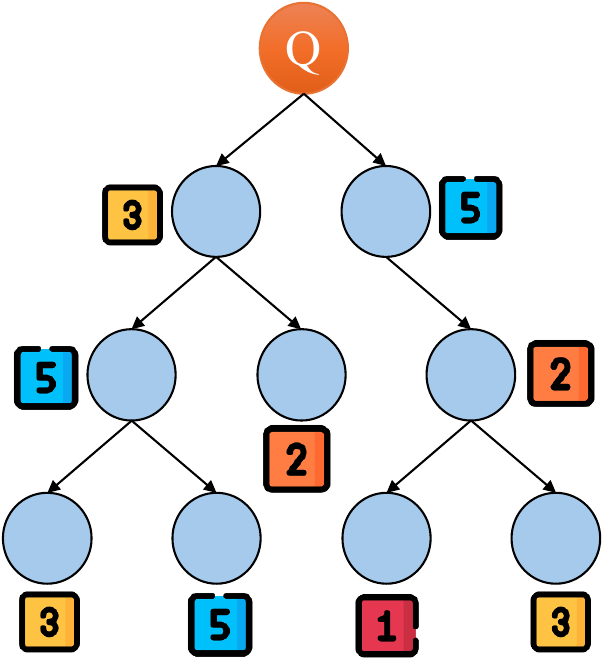}
        \vspace{0.5em}
        (c) RLHF
    \end{minipage}
    \caption{A verifier engineering perspective on SFT, DPO, and RLHF: gray nodes represent sample paths not used in training, while non-gray nodes represent sample paths actively used in the training process.}
    \label{fig:perspective}
\end{figure}

SFT generates candidate responses by employing a linear search strategy that adheres to a singular search path defined by its training data. 
Throughout this process, the verifier classifies each token along the search path as a positive signal, while treating all other tokens as negatives. Subsequently, the foundational model is optimized through imitation learning using a cross-entropy loss function.

Similarly, DPO employs a linear search strategy, maintaining only two distinct search paths: one corresponding to the ``chosen'' data derived from preference pairs and the other to the ``rejected'' data. 
The verifier treats the path associated with the chosen data as positive signals and the path associated with the rejected data as negative signals. 
Subsequently, the foundational model is optimized through the application of pairwise loss functions.

RLHF employs a linear search strategy during the search stage, which is further enhanced by adjusting parameters such as Top-p and temperature to promote exploration within the candidate response generation process. 
As depicted in Figure~\ref{fig:perspective}(c), after search, a value function is utilized to assign scores to each state within the generation trajectory. 
This scoring mechanism estimates the expected return, thereby informing and guiding the optimization process through PPO algorithm. 

By mapping these methods to the stages of verifier engineering, we gain a clearer understanding of how each approach in the search, verify, and feedback stage enhances foundation model capabilities.

\begin{table*}[htbp]
  \centering
    \resizebox{\linewidth}{!}{
    \begin{tabular}{>{\raggedright\arraybackslash}p{7cm}>{\raggedright\arraybackslash}p{2cm}>{\raggedright\arraybackslash}p{5cm}>{\raggedright\arraybackslash}p{4cm}>{\raggedright\arraybackslash}p{2cm}}
    \toprule
    \rowcolor{gray!20}
          & \textbf{Search} & \textbf{Verify} & \textbf{Feedback} & \textbf{Task} \\
    \midrule
    
    \begin{tabular}{@{}l@{}}
    STar~\cite{NEURIPS2022_639a9a17} \\
    RFT~\cite{yuan2023scalingrelationshiplearningmathematical}
    \end{tabular}
    & Linear & Golden Annotation & Imitation Learning & Math \\
    \rowcolor{gray!20}
    CAG~\cite{pan2024contextsequalteachingllms}   & Linear & Golden Annotation & Imitation Learning & RAG \\
    Self-Instruct~\cite{wang-etal-2023-self-instruct} & Linear & Rule-based & Imitation Learning & General \\
    \rowcolor{gray!20}
    \begin{tabular}{@{}l@{}}
    Code Alpaca~\cite{codealpaca} \\
    WizardCoder~\cite{luo2024wizardcoder}
    \end{tabular}
    & Linear & Rule-based & Imitation Learning & Code \\
    ILF-Code~\cite{chen2024improvingcodegenerationtraining}  & Linear & Rule-based \& Code interpreter  & Imitation Learning & Code \\
    \rowcolor{gray!20}
    \begin{tabular}{@{}l@{}}
    RAFT~\cite{dong2023raft} \\
    RRHF~\cite{yuan2023rrhf}
    \end{tabular}
    & Linear & ORM   & Imitation Learning & General \\
    SSO~\cite{xiang2024aligninglargelanguagemodels}   & Linear & Rule-based & Preference Learning & Alignment \\
    \rowcolor{gray!20}
    CodeUltraFeedback~\cite{weyssow2024codeultrafeedback} & Linear & Language Model & Preference Learning & Code \\
    Self-Rewarding~\cite{yuan2024self} & Linear & Language Model & Preference Learning & Alignment \\
    \rowcolor{gray!20}
    StructRAG~\cite{li2024structragboostingknowledgeintensive} & Linear & Language Model & Preference Learning & RAG \\

    LLAMA-BERRY~\cite{zhang2024llamaberrypairwiseoptimizationo1like} & Tree  & ORM   & Preference Learning & Reasoning \\
    \rowcolor{gray!20}
    Math-Shepherd~\cite{wang2024mathshepherdverifyreinforcellms} & Linear & Golden Annotation \& Rule-based & Reinforcement Learning & Math \\
    \begin{tabular}{@{}l@{}}
    RLTF~\cite{liu2023rltfreinforcementlearningunit} \\
    PPOCoder~\cite{shojaee2023executionbasedcodegenerationusing}
    \end{tabular}
    & Linear & Code Interpreter & Reinforcement Learning & Code \\
    \rowcolor{gray!20}
    RLAIF~\cite{lee2023rlaif}  & Linear & Language Model & Reinforcement Learning & General \\
    SIRLC~\cite{pang2023languagemodelselfimprovementreinforcement}  & Linear & Language Model & Reinforcement Learning & Reasoning \\
    \rowcolor{gray!20}
    RLFH~\cite{wen2024onpolicyfinegrainedknowledgefeedback}  & Linear & Language Model & Reinforcement Learning & Knowledge \\
    RLHF~\cite{ouyang2022training} & Linear & ORM   & Reinforcement Learning & Alignment \\
    \rowcolor{gray!20}
    Quark~\cite{lu2022quark}  & Linear & Tool  & Reinforcement Learning & Alignment \\
    ReST-MCTS~\cite{zhang2024restmctsllmselftrainingprocess} & Tree  & Language Model & Reinforcement Learning & Math \\
    \rowcolor{gray!20}
    CRITIC~\cite{gou2024criticlargelanguagemodels} & Linear & Code Interpreter \& Tool \& Search Engine & Verifier-Aware & Math, Code \& Knowledge \& General \\
    Self-Debug~\cite{chen2023teachinglargelanguagemodels} & Linear & Code Interpreter & Verifier-Aware & Code \\
    \rowcolor{gray!20}
    Self-Refine~\cite{madaan2023selfrefine} & Linear & Language Model & Verifier-Aware & Alignment \\
    ReAct~\cite{yao2022react} & Linear & Search Engine & Verifier-Aware & Knowledge \\
    \rowcolor{gray!20}
    Constrative decoding~\cite{li2023contrastivedecodingopenendedtext} & Linear & Language Model & Verifier-Guided & General \\
    Chain-of-verfication~\cite{dhuliawala2023chainofverificationreduceshallucinationlarge} & Linear & Language Model & Verifier-Guided & Knowledge \\
    \rowcolor{gray!20}
    Inverse Value Learning~\cite{lu2024transferableposttraininginversevalue} & Linear & Language Model & Verifier-Guided & General \\
    PRM~\cite{lightman2023lets} & Linear & PRM   & Verifier-Guided & Math \\
    \rowcolor{gray!20}
    KGR~\cite{guan2023mitigatinglargelanguagemodel}   & Linear & Knowledge Graph & Verifier-Guided & Knowledge \\
    UoT~\cite{hu2024uncertaintythoughtsuncertaintyawareplanning}   & Tree  & Language Model & Verifier-Guided & General \\
    \rowcolor{gray!20}
    ToT~\cite{yao2024tree}   & Tree  & Language Model & Verifier-Guided & Reasoning \\
    \bottomrule
    \end{tabular}%
    }
  \caption{This paper provides a comprehensive exploration of the verifier engineering landscape, breaking it down into three core stages: search, verify, and feedback.}
  \label{tab:exist-work}%
\end{table*}%

\subsection{Discussion on three stages of  Verifier Engineering}
The three stages of verifier engineering play a distinct and critical role in enhancing the capabilities of foundation models.
This discussion delves into current challenges and proposes future research directions, focusing on search efficiency, verifier design, and the effectiveness of feedback methods.

\subsubsection{Advanced Search Methods}
The efficiency and effectiveness of candidate response generation are crucial for model performance. The challenge lies in balancing exploration and exploitation, as well as aligning the search process with specific optimization goals. In this subsection, we first discuss the trade-off between exploration and exploitation, and then discuss how goal-aware search can further optimize the search process.

\paragraph{Balancing Exploration and Exploitation}

The balance between exploration and exploitation is a core issue in the search stage of verifier engineering.
Given an instruction, we need to search for candidate responses. For example, in DPO scenarios, we aim to obtain pair-wise responses with different verification results. Regardless of the objective, how to sample desired responses through the trade-off between exploration and exploitation remains a key challenge.
Exploration refers to exploring new action spaces to obtain verified results with larger variance candidate responses, while exploitation utilizes known effective strategies to obtain responses that meet expected validation results. Exploration helps foundation models better understand the outcomes of different actions, while exploitation can obtain satisfactory responses at a lower cost.
However, recent research~\cite{wilson2024llm,nguyen2024balancing,murthy2024rexrapidexplorationexploitation} suggests that existing methods in the post-training process often rely too heavily on exploitation while neglecting the importance of exploration. This imbalance may cause foundation models to get stuck in local optima, limiting their generalization ability. Therefore, how to better balance these two strategies during training is crucial for improving model performance.

\paragraph{Goal-Aware Search}

As shown in Section~\ref{sec:search}, existing search methods primarily rely on probability search from the next token distribution and adjust the search direction through verifier feedback. 
However, this approach has a fundamental issue: the current search approach often lacks a direct correlation with the optimization goal. 
Consequently, we cannot guide foundation models to generate candidate responses that meet our goals during the search stage and can only rely on post-hoc verification results to filter responses that align with our intentions. 
This delayed verification mechanism significantly reduces search efficiency and increases computational overhead.
To address this challenge, search algorithms like A*~\citep{wang2009finding} fuse the cost of the current node with heuristic estimates of the future cost to the goal. Future work can focus on developing more robust goal estimation and fusion methods.

\subsubsection{Open Questions to Verifier Design}

Designing effective verifiers is pivotal for enhancing the performance and reliability of foundation models. However, several unresolved issues remain regarding the optimal design and integration of verifiers. In this subsection, we first examine the considerations involved in verifier design and the need for systematic evaluation frameworks. Subsequently, we explore the complexities associated with combining multiple verifiers, highlighting the challenges that must be addressed to achieve comprehensive and scalable verification systems.

\paragraph{Verifier Design and Systematic Evaluation}
Designing specific verifiers for different types of instructions is essential.
For instructions with explicit output constraints (such as grammatical format and length limitations), rule-based verifiers should be implemented due to their definitive reliability. 
For question-answering instructions, LLM-based verifiers are typically more effective in evaluating subjective metrics like response fluency and information content. 
However, recent studies~\cite{wen2024rethinking} have revealed only weak correlations between existing verifier evaluation metrics and downstream task performance, highlighting significant limitations in the current verifier evaluation framework. 
Therefore, a systematic evaluation framework is needed to comprehensively assess the effectiveness, applicable scope, and limitations of different types of verifiers. 
Such a framework would not only guide the selection and combination of verifiers but also establish best practices for verifier deployment across various task scenarios.

\paragraph{Challenges of Verifier Combinations}

As mentioned above, it is impossible to obtain effective verification results using a single verifier alone. 
Therefore, integrating multiple verifiers is essential to address the diverse requirements of candidate response evaluation. 
To comprehensively enhance foundation model performance across various task scenarios, an effective verifier combination system must be developed.
Building an effective verifier combination system faces three key challenges:

\begin{itemize}
    \item \textbf{Instruction Coverage}: The verifier combination must be capable of handling various types of instructions to ensure the completeness of the evaluation system. Building a comprehensive verifier framework requires a deep understanding of different task characteristics and evaluation needs, including structured output validation, open-ended question assessment, creative task evaluation, etc.
    \item \textbf{Automatic Routing Mechanism}: Different tasks typically require verifiers with varying forms and granularity, demanding an intelligent verifier routing system. This system needs to analyze instruction characteristics and select appropriate verifier combinations accordingly. Based on the PAC theory discussed in Section 2, the chosen verifier combinations should effectively approximate our ultimate optimization objective.
    \item \textbf{Verification Results Integration Strategy}: When multiple verifiers produce different verification results, a reliable decision-making mechanism is needed. This includes dynamic adjustment of different verifier weights, conflict resolution strategies, and methods for synthesizing final scores. Especially, when there are differences in verification results, factors such as the credibility and task relevance of each verifier need to be considered to make reasonable judgments.
\end{itemize}

\subsubsection{Effectiveness of Feedback Methods}
The effectiveness of feedback methods plays a crucial role in shaping the performance and adaptability of foundation models. In particular, we focus on two key questions: 1) whether the feedback method can accurately and efficiently improve the model's performance, and 2) whether it can generalize effectively to other queries.

\paragraph{Key Factors in Designing Feedback Methods}

Different feedback methods have different impacts on foundation model capabilities. 
When selecting feedback methods, it is essential to carefully balance several key factors.
Firstly, the algorithm should demonstrate sufficient robustness to noise in verification results, ensuring the stability of the feedback process.
Secondly, it is crucial to assess how the feedback algorithm based on specific types of verification results impacts the model. Over-optimizing certain capabilities can undermine the model's fundamental capabilities and overall generalization, potentially leading to performance degradation.
Furthermore, foundation models with different capacities may require distinct optimization approaches. Larger models might benefit from more sophisticated feedback methods, while smaller models may need more conservative feedback methods to prevent capacity saturation. 
It is crucial to find an appropriate balance between these factors while considering the model's inherent capabilities.

\paragraph{Generalization over Queries}

Equipped with reliable verifiers and effective feedback methods, the ideal scenario is to achieve comprehensive improvements in foundation model capabilities through optimization on a limited set of queries. 
This necessitates feedback methods to possess strong cross-query generalization abilities. 
Specifically, when we enhance certain foundation model capabilities through feedback on specific queries, these improved capabilities should effectively transfer and persist when handling novel queries. 
However, generalization also faces significant challenges: different queries may require the model to invoke distinct capabilities, and incorrect generalization might lead to the model inappropriately applying certain capabilities in unsuitable scenarios, potentially degrading performance. 
Therefore, feedback methods must not only promote effective generalization but also prevent over-generalization and incorrect transfer of capabilities.

\subsection{Verifier Engineering Implementation}

In this section, we will discuss the problems we may meet in verifier engineering implementation in different stages to ensure efficiency, scalability, and reliability.

\paragraph{Search}
Achieving high search efficiency is a key objective in verifier engineering. Excessive search often slows down the entire verifier engineering pipeline.

To address this, most current LLM-based PPO algorithms sample only a single response for optimization.
On the other hand, RLHF-like algorithms commonly incorporate importance sampling~\cite{schulman2017proximal,xie2019towards} techniques to enhance search efficiency, minimizing the need for frequent switching between search, verify, and feedback stages while simultaneously improving sample utilization.

\paragraph{Verify}
Verifier efficiency is also a key goal in verifier engineering for giving timely and effective verification results.

When handling multiple instructions from various sources, it's crucial to employ different combinations of verifiers with different abilities to ensure accurate verification results. Determining the optimal approach to deploy all verifiers online and dynamically schedule them each time to minimize resource consumption while maximizing efficiency presents a challenging problem.

Delivering effective verification results involves addressing two major challenges: (1) ensuring the verifier’s knowledge remains synchronized with the policy model, and (2) selecting the optimal verifier combination when capabilities vary or conflict across verifiers. 
For instance, InstructGPT~\cite{ouyang2022training} employs a human-annotated reward model as a verifier, and to counter the limitations of a static verifier, it periodically re-annotates reward model data to align its evaluation capabilities with the evolving policy model outputs.
Furthermore, \citet{quan2024dmoerm} leverages a Mixture of Experts architecture to combine multiple verifiers with different strengths. Experiential Co-Learning~\cite{qian2023experiential} also draws on the knowledge of diverse foundation models to provide more robust verification results.

\paragraph{Feedback}
For highly efficient feedback, it’s essential not only to enhance the feedback algorithm itself but also to optimize the entire workflow.

To increase training and inference efficiency, LoRA~\cite{hu2021lora,xin-etal-2024-beyond} improves training efficiency by reducing the number of trainable parameters and vLLM~\cite{kwon2023efficient} enhance inference efficiency.

To optimize the entire workflow, deciding when to apply feedback methods is crucial~\cite{tang2024understanding}. 
For training-based feedback, understanding the performance gap between online and offline feedback methods is key. Research has shown that online feedback methods can maximize the capabilities of foundation models by providing timely verification results on the on-policy candidate responses, albeit requiring frequent sampling, which can be time-intensive. 
In contrast, offline feedback methods enable comprehensive response exploration by leveraging pre-prepared datasets for training, thereby streamlining the process. However, this approach tends to have lower data utilization efficiency.
This highlights the importance of balancing online and offline feedback methods.
In inference-based feedback, deciding when to call a verifier is essential. \citet{jiang2023active} demonstrate that retrieval based on detecting uncertainty in foundation model internal states enables the model to perform better, suggesting that invoking the verifier on demand yields more effective and efficient outcomes for maximizing foundation model capabilities.

\section{Conclusion}
This paper introduces the concept of verifier engineering and explores the significant shift in research paradigms from feature engineering to data engineering, and ultimately to verifier engineering. 
Our framework provides implications and insights demonstrating that verifier engineering can optimize the capabilities of foundation models through a closed-loop feedback cycle encompassing search, verification, and feedback. 
Furthermore, we categorize existing search algorithms based on their granularity and schemes, review current verifiers, and classify feedback methods from both training-based and inference-based perspectives.
Finally, we discuss the challenges that verifier engineering currently faces. 
Through this paper, we aim to stimulate further discussions and promote practical applications in the field of verifier engineering for achieving Artificial General Intelligence.

\bibliographystyle{icml2024}
\bibliography{example_paper.bib}
\end{document}